# An Extended Beta-Elliptic Model and Fuzzy Elementary Perceptual Codes for Online Multilingual Writer Identification using Deep Neural Network

Thameur Dhieb, Sourour Njah, Houcine Boubaker, Wael Ouarda, Mounir Ben Ayed, and Adel M. Alimi
REGIM-Lab.: REsearch Groups in Intelligent Machines
University of Sfax
National School of Engineers (ENIS), BP 1173, Sfax, 3038, Tunisia
{thameur.dhieb.tn, sourour.njah, houcine-boubaker, wael.ouarda, mounir.benayed, adel.alimi}@ieee.org

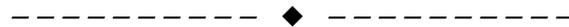

**Abstract**—Actually, the ability to identify the documents' authors provides more chances for using these documents for various purposes. In this paper, we present a new effective biometric writer identification system from online handwriting. The system consists of the preprocessing and the segmentation of online handwriting into a sequence of Beta strokes in a first step. Then, from each stroke, we extract a set of static and dynamic features from new proposed model that we called Extended Beta-Elliptic model and from the Fuzzy Elementary Perceptual Codes. Next, all the segments which are composed of N consecutive strokes are categorized into groups and subgroups according to their position and their geometric characteristics. Finally, Deep Neural Network is used as classifier. Experimental results reveal that the proposed system achieves interesting results as compared to those of the existing writer identification systems on Latin and Arabic scripts.

—————— ◆ ——————

## 1 INTRODUCTION

THERE are typically two categories of biometric modalities: physiological biometrics (e.g. iris, fingerprint, retinal, face) [1], [2], [3] and behavioral biometrics (e.g. signature, voice, Rhythm of typing keys , gait) [4], [5], [6]. Writer identification therefore belongs to the category of behavioral biometrics.

Writer identification aims to determine the writer of a document among a list of authors according to the similarity or distance with respect to their handwritings [7], [8]. Currently, the identification of individuals based on handwriting is an active area where researchers are trying various methods and different approaches to increase identification rates. This is due to its feasibility in a wide range of applications like forensic science, control access, digital rights management and financial transactions [9], [10], [11], [12].

Moreover, writer identification approaches are divided into two broad categories: offline and online [13], [14], [15]. In offline category, the input represents spatial data using an image. In online category, the input represents temporal data using a list of points representing the trajectory of handwriting. In another classification manner, writer identification methods are also categorized into two main types: text dependent and text independent [16], [17], [18]. Text dependent methods are dependent on a given text content, while text independent methods are not limited to a specific text content in order to identify writers. The proposed system in this paper falls into the online category with text independent method as shown in Figure 1.

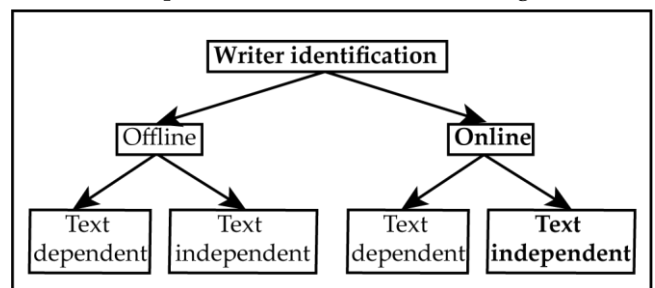

Fig. 1. Writer identification categories

Indeed, the writer identification task faces several challenges and difficulties. One of the main difficulties is the intra-writer variability factor [19]. In this respect, the psychological characteristics of a person (e.g. apprehension, education, fright and stress level) and the neurophysiological characteristics (e.g. nervous system, hand, muscles, arm, and finger) influence the handwriting style of the writer. Accordingly, a writer's handwriting can unconsciously modify during his lifetime [20].

Another difficulty to the writer identification task is the inter-writer variability factor [21]. On the first hand, all the writers try to write the same characters in the same way.

On the second hand, however, the variations of writing styles among the different writers should exceed the intra-writer variations for every single writer [22] in order to distinguish different writers.

Consequently, an important challenge associated with the writer identification field is to define a set of features able to characterize the different samples of handwriting. These samples are not usually stable and show a vast variability from the same writer over time, or from different writers. Hence, it may be necessary to enhance or introduce new features extraction techniques able to well characterize the handwriting of a person.

According to other researchers, the dynamic features (e.g. velocity, time) and the static features (e.g. character structure, shape) have been found representative to characterize the handwriting style [23], [24].

Therefore, in this paper, we develop an online text independent writer identification system based on new approach inspired from Beta-Elliptic model [25], [26], [27] and Fuzzy Elementary Perceptual Codes [28], [29]. Our choice is justified by the fact that the Beta-Elliptic model allows to extract static and dynamic features. In addition, we are interested in taking advantage of the representations of handwriting by perceptual codes.

This paper is organized as follows: Section 2 discusses related works in online text independent writer identification systems. Section 3 describes our ongoing research for the development of an online writer identification system including the definition of new model named Extended Beta-Elliptic model and Fuzzy Elementary Perceptual Codes. The experiments and the obtained results are presented and discussed in section 4, while section 5 concludes this paper with a summary and an outlook towards future works.

## 2 RELATED WORKS

In recent years, remarkable progress has been made and several approaches have been adopted in addressing the problem of online text independent writer identification. Among these, we can cite the system of [30]. In this system, different sets of features are extracted from the acquired data which are point-based feature set, stroke-based feature set, extended point-based feature set, off-line point-based feature set and all point-based feature set. These features used to train Gaussian mixture models. The training data of all writers are used to train a Universal Background Model.

To extract features from online handwriting, Li *et al.* [31] propose shape codes to characterize the trajectory direction and temporal sequence codes to characterize the pressure and the speed of handwriting. To identify the writer, they implement decision and fusion strategy.
Shivram *et al.* [32] model writing styles as a shared component of an individual's handwriting by using a three level hierarchical Bayesian structure called Latent Dirichlet allocation. These writing styles are trained with n class Support Vector Machines where n represents the number of writers.

TABLE 1
A SUMMARY OF SOME ONLINE TEXT INDEPENDENT WRITER IDENTIFICATION SYSTEMS

| System | Features | Classifier |
|---|---|---|
| [30] | Point-based feature set, stroke-based feature set, extended point-based feature set, off-line point-based feature set and all point-based feature set. | Gaussian Mixture Model (GMM) |
| [31] | Shape codes and temporal sequence | Nearest neighbor Classification |
| [32] | Probability distribution feature | Support vector machines |
| [33] | Higher-Order United Moment Invariant | Decision Trees, k-Nearest Neighbors, Sequential Minimal Optimization, Multilayer Perceptrons and Naive Bayes |
| [34] | Feature space models and writer style space models | Support Vector Machines. |
| [36] | Speed and the coordinates of trajectories | Nearest prototype method, modified tf–idf approach |
| [37] | Path-signature feature | Deep convolutional neural network |
| [38] | Codebook descriptors | Support Vector Machines |

Jalil *et al.* [33] propose an approach of online text independent writer identification following four steps: features extraction using Higher Order United Moment Invariant, features ranking based on Grey Relational Analysis, discretization, and lastly classification using five classifiers which are Decision Trees, k-Nearest Neighbors, Sequential Minimal Optimization, Multilayer Perceptrons and Naive Bayes.

In another works, Shivram *et al.* [34] present a comparative study of two approaches to online text independent writer identification, namely feature space models and writer style space models. For classification, they use Support Vector Machines. Experiments show that writer style space models give higher performance than feature space models with 90.02% of writer identification rate for 43 writers from IBM_UB_1 dataset [32], [35].

Singh *et al.* [36] extract the coordinates x, y and the speed feature of the different strokes which are segmented into sub-strokes. Then, they employ an unsupervised learning scheme called subtractive clustering in order to derive the specific sub-strokes of a given writer and to model his writing styles. Finally, they propose a modified scoring scheme for the identification process.

Yang *et al.* [37] propose an end-to-end system for text independent writer identification from online handwriting. They do the preprocessing step. Then, they introduce DropSegment in order to achieve data augmentation. Besides, they define path signature feature maps so that they can improve the performance of their system. Finally, they use deep convolutional neural network so as to exploit the spatial sparsity of handwriting.

Venugopal *et al.* [38] propose an online writer identification framework. For realizing their framework, they derive a strategy that encodes the sequence of feature vectors extracted at sample points of the temporal trace with descriptors obtained from a codebook to improve the Vector of Local Aggregate Descriptor (VLAD). For classification, they use Support Vector Machines. Recently, the authors improve theier codebook descriptors in [39]. In fact, given a codebook of size k, they consider the descriptors of only k−1 codevectors and they extract point-based features by incorporating a gap parameter to build the final descriptor.

A summary of some online text independent writer identification systems are presented in Table 1.

Based on the fact that the objective of all writer identification systems is to authenticate the identity of writers, the most important challenge in this task is the definition of features able to well characterize the writer. Consequently, we define in this paper an extended version of Beta-Elliptic model and Fuzzy Elementary Perceptual Codes for the purpose of differentiating the styles of handwriting.

## 3 ARCHITECTURE OF THE PROPOSED SYSTEM

The proposed system has five major steps: Preprocessing and segmentation of handwriting into Beta strokes, features extraction, handwritng segments pre-classification, Deep Neural Network training and identification phase. Figure 2 provides an overview of our proposal.

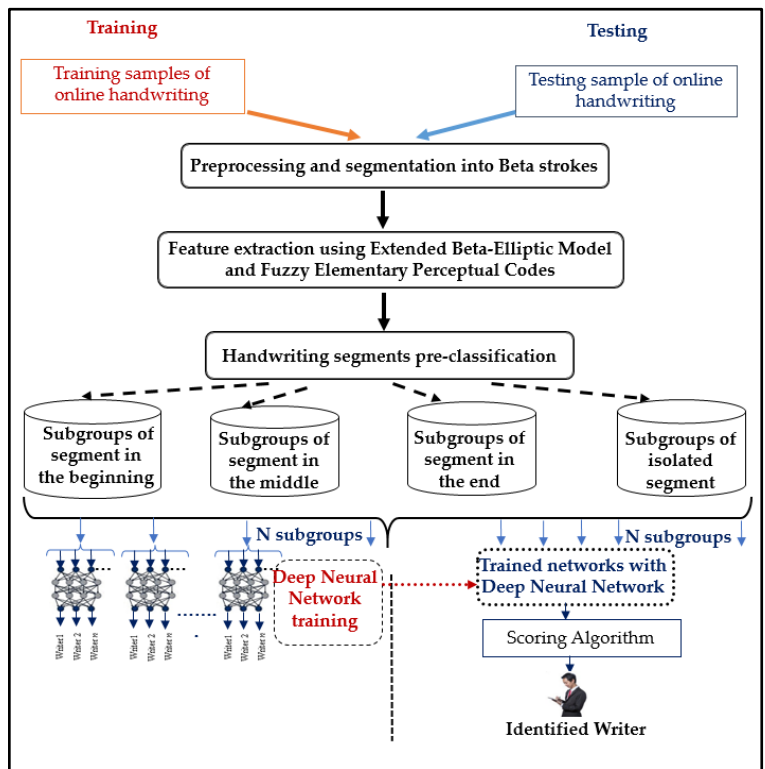

Fig. 2. The general architecture of the proposed system

The first step aims to process the online handwriting and decompose it into Beta strokes which are the result of a superimposition of time-overlapped velocity profiles. The features extraction step aims to model the handwriting trajectories by discriminative features allowing to characterize the different writers styles. The writers styles are examined through sliding window composed of N successive Beta strokes called segment. In fact, All the obtained segments will be pre-classified into groups and subgroups, as detailed in section 3.3, before being assigned to a given writer using Deep Neural Network algorithm. The last step collect the output rates of the neural network classification for a set of successive segments composing the handwritten script to determine its writer's identity.

### 3.1 Preprocessing and segmentation

Preprocessing step is the first and essential step to handwriting processing operations such as normalization and smoothing in order to remove the irregularities of handwriting. In our system, a low-pass filtering is involved. It is a Chebyshev type II filter with a cut-off frequency of $f_{cut}$ =12Hz accorded with handwriting frequency ripple on the input path to mitigate the effect of noise and errors due to temporal and spatial quantification [40]. Then, we normalize the size of the handwiting to adjust its height to a fixed value h = 128 keeping the same ratio length / height. Knowing that the inputs of online handwriting are the coordinates of the trajectory of pen points represented as a function of time x(t) and y(t), the segmentation problem of handwriting is considered as a challenging task. It consists in the different operations that must be performed to get the basic units of the handwriting such as strokes or graphemes [41] [42].

To segment online handwriting, we split the input sig-nal that represents the velocity profile of a handwriting trajectory in smaller units that we called Beta strokes by sliding the online handwriting. In fact, choosing a Beta function to model the curvilinear velocity profile $V\sigma(t)$ which represents the resulting response to the finished impulses. It can be calculated through the following formula:

$$V_\sigma(t) = \sqrt{\left(\frac{dx(t)}{dt}\right)^2 + \left(\frac{dy(t)}{dt}\right)^2} \quad (1)$$

Many researchers use different points to split the pen-path into smaller entities. Commonly used significant points are local extrema in curvature [43], local extrema in vertical or horizontal direction [44], local extrema in velocity [45] and points inflection [46] [47] [48].

In our proposed model, the system will proceed to segment the velocity profile and the handwriting trajectory in concatenated Beta strokes limited between successive velocity local minimums or double inflexion points as shown in Figure 3.

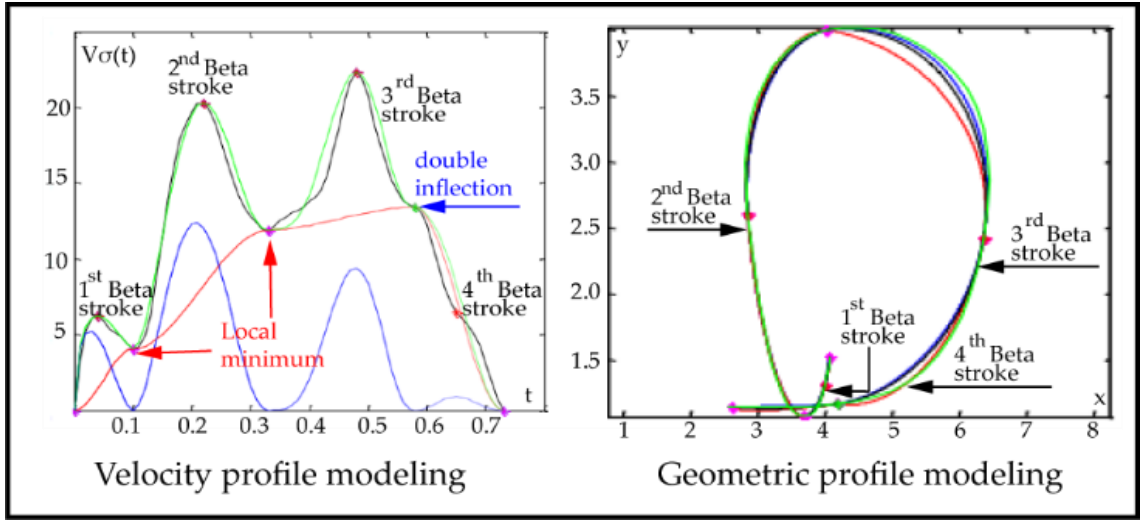

Fig. 3. Example of segmentation in Beta strokes

## 3.2 Features extraction

The purpose of features extraction in writer identification is to obtain specific parameters from handwriting that are able to distinguish the handwriting of one writer from other writers [49]. In this study, we propose a new model that we called Extended Beta-Elliptic model and we use the Fuzzy Elementary Perceptual Codes to characterize the handwriting.

### 3.2.1 Beta-Elliptic model

Founded on the hypothesis that handwriting trajectory represents a movement learned and planned in advance, the Beta-Elliptic model is based on modelling the speed profile of the movement by a Beta function and by using the ellipse to model the shape trajectory.

Each stroke corresponds in the kinematics profile to the generation of one Beta signal as shown in (2):

$$pulse\, \beta(K,t,q,p,t_0,t_1) = \begin{cases} K.\left(\dfrac{t-t_0}{t_c-t_0}\right)^p \cdot \left(\dfrac{t_1-t}{t_1-t_c}\right)^q & if\ t \in [t_0,t_1] \\ 0, & elsewhere \end{cases} \quad (2)$$

with:
- $t_0$ is the starting time of generated impulse.
- $t_1$ is the ending time of generated impulse.
- $t_c$ is the instant when the generated Beta function reaches its maximum value K.
- K is the Beta impulse amplitude.
- p, q are intermediate parameters that have an influence on the width and the symmetry of the generated Beta shape. p, q $\in$ IR.

Figure 4 shows an example of a shape of symmetrical generated Beta function.

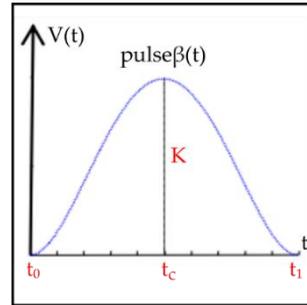

Fig. 4. Shape of symmetrical generated Beta function.

Different shapes of Beta function displayed in Figure 5 according to the values of its refinement and asymmetry parameters p and q with a same amplitude K = 1.

Thus, the generation of a velocity model for an online handwriting is the result of the algebraic sum of Beta profiles as defined in (3):

$$V(t) = \sum_{i=1}^{n} V_i(t-t_{0i}) = \sum_{i=1}^{n} pulse\, \beta_i(K_i,t,q_i,p_i,t_{0i},t_{1i}) \quad (3)$$

In the static profile, each elementary movement, named stroke, is executed in the space domain from an arbitrary starting position and checks a monotony curvature variation that can be assimilated to an elliptic arc.

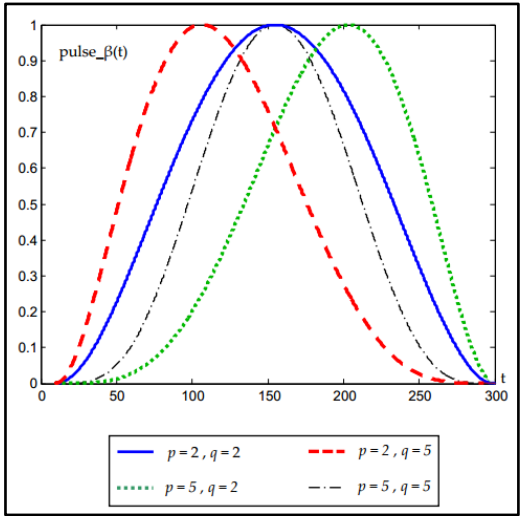

Fig. 5. Different forms of Beta function according to the values of parameters p and q

### 3.2.2 Features extraction performed by the Extended Beta-Elliptic model

We consider in our proposed model that the handwriting velocity modeling superposes the successive velocity Beta impulses to a component of continuous drag developed by training. Thus, the time axis of the velocity profile is decomposed into intervals that represent cycles of acceleration, deceleration and braking. Each time interval T=[$t_0$, $t_1$] is limited by a successive local minimums or double inflexion points of velocity: $V_i=V_o(t_0)$ and $V_f = V_o(t_1)$. During each interval, the curvilinear velocity (shown in Figure 6.a) can be divided into two components:

- The first is an impulsive component $V_{Imp}(t)$ that represents a velocity impulsion with finished energy during the time interval T, engendered by a cycle of acceleration, deceleration and braking. The impulsive component can be represented by a generated impulse as defined in (4):

$$V_{Imp}(t) = K \cdot \left(\frac{t-t_0}{t_C-t_0}\right)^p \cdot \left(\frac{t_1-t}{t_1-t_C}\right)^q \quad (4)$$

with:
- $t_0$ is the starting time of Beta function.
- $t_1$ is the ending time of Beta function.
- $t_c$ is the instant when the Beta function reaches its maximum value K.
- K is the Beta impulse amplitude.

- The second is a continuous training component, which engenders the energy that permits the continuous passage from a trajectory segment to another separated by a local minimum of curvature radius. Thus, the variation tuning to the time of the continuous training component is defined in (5) by a monotonous polynomial function of third degree:

$$V_{Tra}(t) = A \cdot \left[\frac{(t-t_0)^3}{3} - \frac{(t_1-t_0)\cdot(t-t_0)^2}{2}\right] + V_i \quad (5)$$

where 
$$A = -6 \cdot \frac{V_f - V_i}{(t_1-t_0)^3} \quad (6)$$

- $t_0$ is the starting time of Beta function.
- $t_1$ is the ending time of Beta function.
- $V_i$ is the velocity at the starting time of Beta function.
- $V_f$ is the velocity at the ending time of Beta function.

The reconstituted curvilinear speed of tracing is obtained by the sum of its impulsive component together with the continuous training component as defined in (7):

$$V_R(t) = V_{Imp}(t) + V_{Tra}(t) \quad (7)$$

From another angle, in the static profile, we adopt a hypothesis of simplification for trajectory modeling by elliptic arcs as shown in Figure 6.b. In fact, given that in our proposed model, Beta strokes stretch along intervals ranging from a local velocity minimum to the next velocity minimum or double inflexion. They are modeled by a couple of elliptic arcs $E_1$ ($a_1$, $b_1$, $\theta_1$, $\theta_{p1}$) and $E_2$ ($a_2$, $b_2$, $\theta_2$, $\theta_{p2}$) sharing the same major axis direction ($\theta_1 = \theta_2 = \theta$) and checking the continuity of the curvature function at the link point $M_2$ as shown in (8).

$$a_2 = a_1 \cdot \sqrt{\frac{b_2}{b_1}} \quad (8)$$

This last condition leads to a relation between the two arcs minor and major axis lengths allowing to resume the parametric model. Then, trajectory stroke covered by this two tied elliptic arcs can be described by only six parameters ($a_1$, $b_1$, $b_2$, $\theta_{p1}$, $\theta$, and $\theta_{p2}$) as represented in Figure 7.

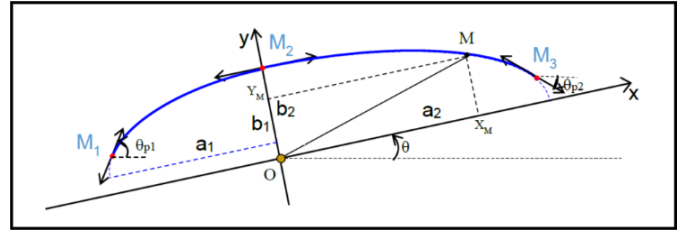

Fig. 7. Explanation of geometric features generated by the Extended Beta-Elliptic model

The retained number of Beta strokes per segment in the proposed model which maximizes the identification rate is N=2. Thus, a vector of 14 features is obtained for each stroke as detailed in Table 2.

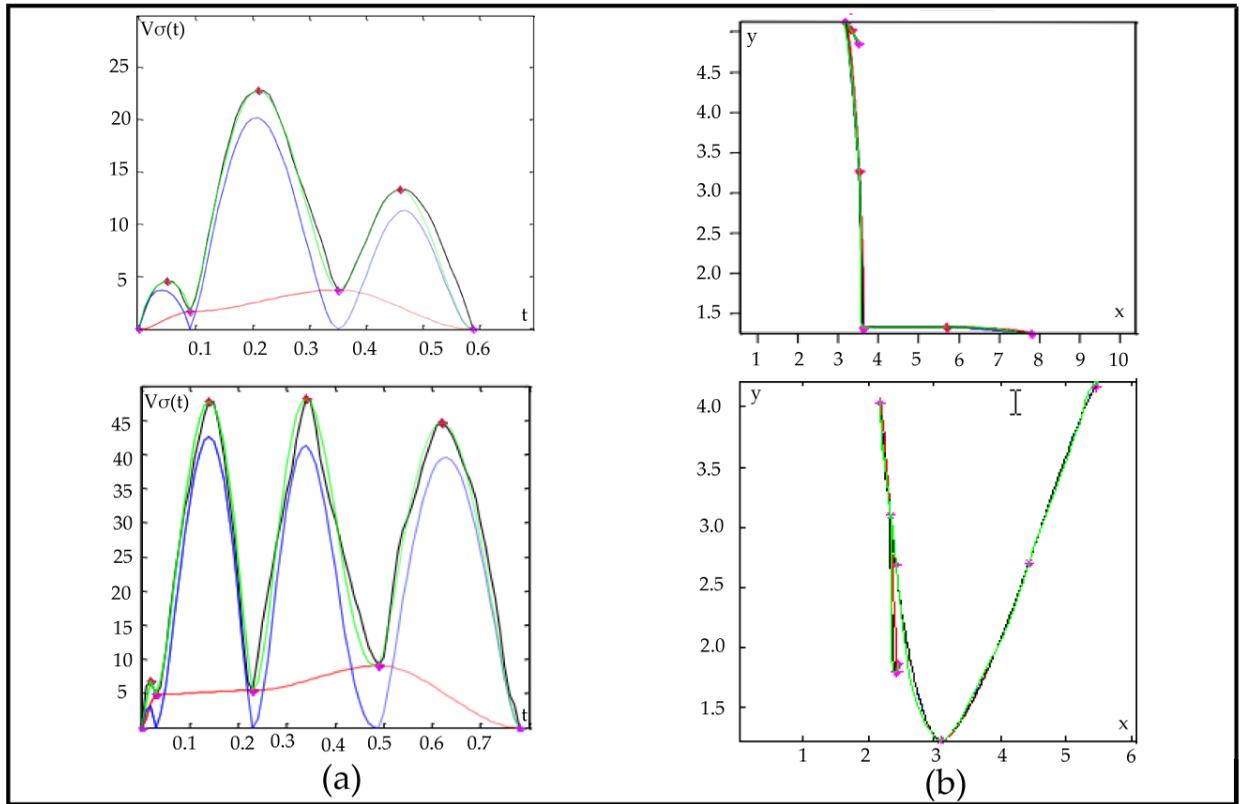
Fig. 6. Application of the Extended Beta-Elliptic model on handwriting trajectory of 'L' and 'V' characters:
(a) Velocity profile modeling and (b) Geometric profile modeling

TABLE 2
FEATURES EXTRACTION GENERATED BY THE EXTENDED BETA-ELLIPTIC MODEL

| Feature | Parameter and formula | Signification |
|---|---|---|
| $f1$ | $\Delta t = (t_1 - t_0)$ | Beta impulse duration |
| $f2$ | $RapT_c = \dfrac{t_c - t_0}{\Delta t}$ | Beta impulse asymmetry report or culminating time |
| $f3$ | $P$ | Beta shape parameter of the neuro-muscular impulse |
| $f4$ | $K$ | Beta impulse amplitude |
| $f5$ | $V_i$ | Initial training velocity amplitude of the current stroke |
| $f6$ | $V_{fin}$ | Final training velocity amplitude of the current stroke |
| $f7$ | $\dfrac{k_i}{training}$ | Beta impulse amplitude report respect to medium value of the training component |
| $f8$ | $a_1$ | Major axis half length of the ellipses supporting the first arc |
| $f9$ | $b_1$ | Half length of small axis of the ellipse including the first arc |
| $f10$ | $b_2$ | Half length of small axis of the ellipse including the second arc |
| $f11$ | $\theta_{P1}$ | Inclination angle of the tangents at the stroke endpoint M1. |
| $f12$ | $\theta$ | Ellipse major axis inclination angle |
| $f13$ | $\theta_{P2}$ | Inclination angle of the tangents at the stroke endpoint M3 |
| $f14$ | POS_STROKE | Stroke position |

The first seven features help us to discriminate between the writers in the speed of handwriting. In fact, these features aims to capture the dynamic aspects of the writing behavior of an individual.

The last seven features help us to discriminate between the writers in the way of writing. In other words, these features aims to capture the geometric characteristics present in the handwriting to identify the writer.

### 3.2.3 Fuzzy Elementary Perceptual Codes (FEPC)

Fuzzy Elementary Perceptual Codes consisting in allocating an elementary perceptual code for each stroke with a certain membership degree. According to the hypothesis that handwriting is a sequence of perceptual codes, we use four elementary perceptual codes presented in Table 3.

TABLE 3
THE ELEMENTARY PERCEPTUAL CODES

| N° | Elementary Perceptual Code (EPC) | Shape |
|---|---|---|
| 1 | $EPC_1$ : Valley | — |
| 2 | $EPC_2$ : Left oblique shaft | / |
| 3 | $EPC_3$ : Shaft | \| |
| 4 | $EPC_4$ : Right oblique shaft | \ |

In order to identify the different EPCs, we use the ellipse major axis inclination angle $\theta$ extracted of each stroke presented in our proposed model. The trigonometric circle is segmented into eight equidistant regions of θ and all corresponding EPCs, as shown in Figure 8.

Between two successive regions of θ, we consider an overlapped interval values noted cst equal to π/16 where we have a problem of indecision about the corresponding EPC. Figure 9 shows an example of the perceptual problem decision about the belongings of θ example presented in Figure 8 which is equal to (4.83/8)π.

Knowing that we use the concept of fuzzy logic sets theory, we take the value example of θ equal to (4.83/8)π, the corresponding $EPC_s$ are: $EPC_3$ 'Shaft' with 67% of membership degrees, and $EPC_4$ 'Right oblique shaft' with 33% of membership degrees.

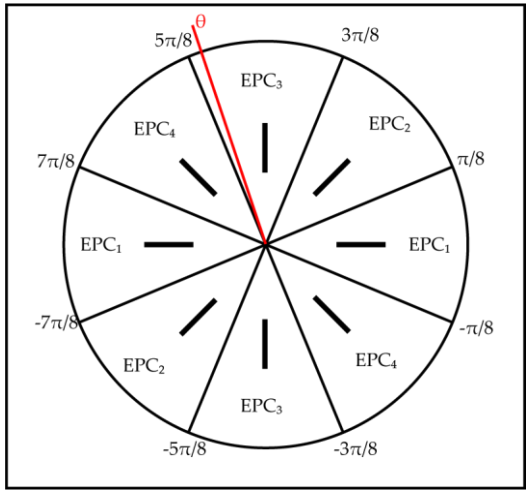

Fig. 8. Deviation angles regions and EPCs on the trigonometric circle

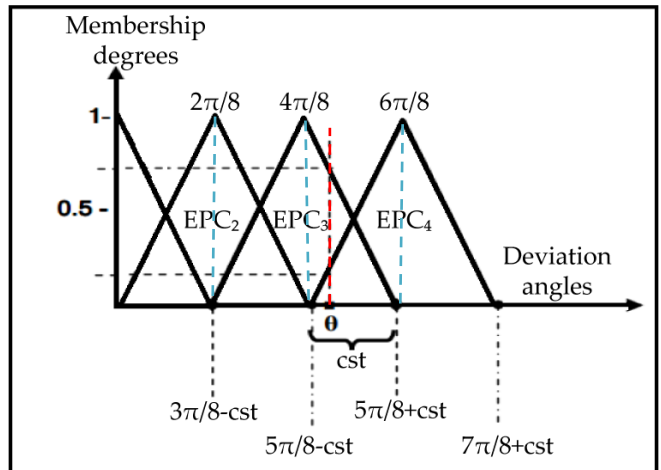

Fig. 9. Example of perceptual problem decision using fuzzy logic

Thus, a vector of 4 features is obtained for each stroke as detailed in Table 4.

TABLE 4
FEATURES EXTRACTION GENERATED BY THE FUZZY ELEMENTARY PERCEPTUAL CODES

| Feature | Parameter and formula | Signification |
|---|---|---|
| $f1$ | $EPC_1$ | Membership degree of $EPC_1$ |
| $f2$ | $EPC_2$ | Membership degree of $EPC_2$ |
| $f3$ | $EPC_3$ | Membership degree of $EPC_3$ |
| $f4$ | $EPC_4$ | Membership degree of $EPC_4$ |

These features can contribute in discriminating between the writers. To prouve this hypothese, we combine these features with the Beta-Elliptic model and with the Extended Beta-Elliptic model. The results presented in section 4.

### 3.3 Handwriting segments pre-classification

The developed algorithm considers a window composed of N successive Beta strokes to constitute a handwriting segment sliding on the pen trajectory. It categorizes each obtained handwriting segment according to its position in word, in one of the four following groups:
1) Segments in the beginning;
2) Segments in the middle;
3) Segments in the end;
4) Isolated segments.

Each one of the four groups is divided into sub groups of trajectory segments sharing the same visual shape by examination of:

1) the starting point, the arrival point and the points extremas in the four positions (left, right, top, low);

2) the different vertical position of trajectory segment;

3) the evolution of the geometric characteristics of the trajectory and in particular its tangent direction along the segment path.

Therefore, for Latin and Arabic script, we have constituted:

- 10 subgroups from the group segment in the beginning as shown in Table 5.

- 12 subgroups from the group segment in the middle as shown in Table 6.

- 10 subgroups from the group segment in the end as shown in Table 7.

- 8 subgroups from the group isolated segment as shown in Table 8.

### 3.4 Deep Neural Network training

In the training, we use stacked sparse autoencoders plus softmax classifier. Stacked sparse autoencoder, as mentioned is an unsupervised learning algorithm that perform a back propagation in order to tune the target values to be equal to the inputs. In fact, stacked sparse autoencoders architecture have a series of inputs, outputs and hidden layers. Starting from the first hidden layer, there is always reconstructing of the output of the previous layer. As for softmax classifier, it is a supervised model used for training with the labels of the training data. Figure 10 presents an overview on the proposed general architecture of Deep Neural Network training. For simplicity and clarity, we do not expose the decoder in this Figure.

TABLE 5

SUBGROUPS OF SCRIPT SEGMENTS IN THE BEGINNING

| N° | Subgroup | Segment |
|---|---|---|
| 1 | opened right curve | |
| 2 | opened left curve | |
| 3 | beginning ascending shaft | |
| 4 | beginning descending shaft | |
| 5 | half shaft in the beginning | |
| 6 | broad occlusion in the beginning | |
| 7 | occlusion e beginning for Latin script | |
| 7 | nabra in the beginning for Arabic script | |
| 8 | residual segments in the low | |
| 9 | residual segments in the top | |
| 10 | residual segments in the median | |

TABLE 6

SUBGROUPS OF SEGMENTS IN THE MIDDLE

| N° | Subgroup | Segment |
|---|---|---|
| 1 | medium ascending shaft | |
| 2 | medium descending shaft | |
| 3 | broad occlusion in the medium | |
| 4 | begin of occlusion in the medium | |
| 5 | medium occlusion | |
| 6 | half shaft in the medium | |
| 7 | e in the medium for Latin script | |
| 7 | narrow occlusion in the medium for Arabic script | |
| 8 | n in the medium for Latin script | |
| 8 | middle half Nabra for Arabic script | |
| 9 | medium valley for Latin script | |
| 9 | ligature madda | |
| 10 | residual segments in the low | |
| 11 | residual segments in the top | |
| 12 | residual segments in the median | |

TABLE 8

SUBGROUPS OF LATIN SCRIPT ISOLATED SEGMENTS

| N° | Subgroup | Segment |
|---|---|---|
| 1 | isolated shaft | |
| 2 | isolated begin of occlusion | |
| 3 | isolated pocket | |
| 4 | isolated j for Latin script | |
| 4 | diacritics for Arabic script | |
| 5 | isolated O occlusion for Latin script | |
| 5 | curvy leg isolated for Arabic script | |
| 6 | residual segments in the low | |
| 7 | residual segments in the top | |
| 8 | residual segments in the median | |

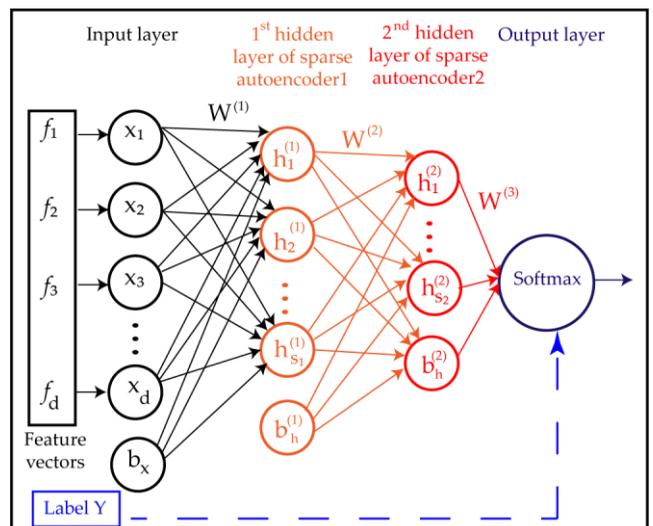

Fig. 10. General architecture of a Deep Neural Network training

TABLE 7
SUBGROUPS OF LATIN SCRIPT SEGMENTS IN THE END

| N° | Subgroup | Segment |
|---|---|---|
| 1 | end ascending shaft | |
| 2 | end descending shaft | |
| 3 | half shaft in the end | |
| 4 | curvy leg end | |
| 5 | end occlusion | |
| 6 | e in the end for Latin script | |
| 6 | narrow occlusion in the end for Arabic script | |
| 7 | pocket leg end for Latin script | |
| 7 | Nabra in the end for Arabic script | |
| 8 | residual segments in the low | |
| 9 | residual segments in the top | |
| 10 | residual segments in the median | |

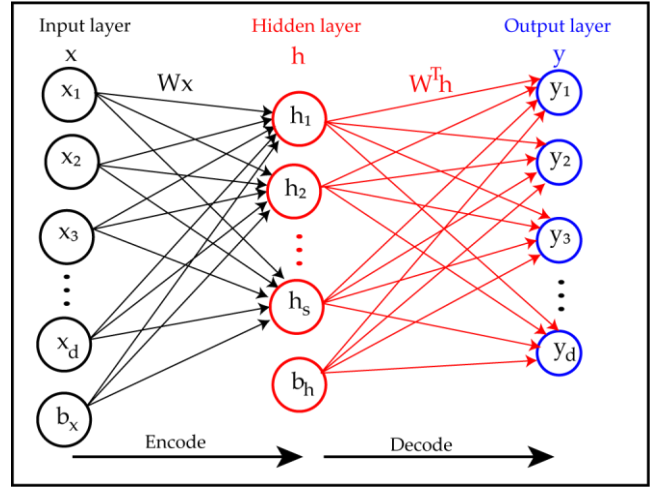

Fig. 11. The basic structure of sparse autoencoder neural network using *d* input units and *s* hidden units

The simplest sparse autoencoder consists in a single hidden layer, *h*, that is connected to the input vector *x* by the learning parameters *W* forming the encoding step. The hidden layer then outputs to a reconstruction vector *y*, using a tied weight matrix $W^T$ to form the decoder. Thus, we represent *h* and *y* respectively in (9) and (10):

$$h = f(Wx + b_x) \quad (9)$$

$$y = f(W^T h + b_h) \quad (10)$$

where:
*f*: activation function.
$b_x$: bias vectors for the hidden layers.
$b_h$: bias vectors for the output layers.

The sparse autoencoder contains three types of layer: First layer used as encoder, last layer used as decoder and a hidden layer. The first layer tries to map the input features vector into the hidden layer transformation. That way, the input and the output vectors have the same dimension. The basic structure of sparse autoencoder neural network used in our proposed general architecture is presented in Figure 11.

The training operation is realized via the following four steps:
1) A first sparse autoencoder is applied to the vectors features extracted x to learn a primary representation $h^{(1)}(x)$ by adjusting the weight $W^{(1)}$;
2) This primary representation $h^{(1)}(x)$ is treated as input to a second sparse autoencoder to learn the secondary representation $h^{(2)}(x)$ by adjusting the weight $W^{(2)}$;
3) This secondary representation $h^{(2)}(x)$ is fed into a softmax classifier, and it is trained to map $h^{(2)}(x)$ using label Y by adjusting the weight $W^{(3)}$;
4) Finally, we stack the encoders from the sparse autoencoders together with the softmax classifier layer to form our Deep Neural Network.

The number of input units in the input layer is defined according to the features extraction techniques used. In fact, the training leads to the segment containing N=2 Beta strokes. So, a vector of 28 features (2 x 14 features) and a vector of 36 features (2 x [14 +4 features]) are used respectively for the Extended Beta-Elliptic model and for the combinaison between the Extended Beta-Elliptic model and the Fuzzy Elementary Perceptual Codes in the input.

The first and second hidden layers have $d_{h(1)} = s_1 = 240$ and $d_{h(2)} = s_2 = 120$ hidden units respectively. All parameters used in the sparse autoencoders training are presented in Table 9. The softmax layer is of 43 outputs that represent the class number (the same as number of writers in our experiments).

TABLE 9
PARAMETERS USED IN SPARSE AUTOENCODERS TRAINING

| Parameter | First sparse autoencoder | Second sparse autoencoder |
|---|---|---|
| Hidden Size | 240 | 120 |
| Max Epochs | 400 | 100 |
| L2 Weight Regularization | 0.004 | 0.002 |
| Sparsity Regularization | 4 | 4 |
| Sparsity Proportion | 0.15 | 0.1 |

## 3.5 Identification process

In order to evaluate the performance of our system, we extract the features from the testing samples just as we did in the training phase. After that, we calculate the sum of the simulation output of the correspondent trained networks with Deep Neural Network for all the subgroups. Then we rank the writers according to their obtained sum of output affectation rate $T_{affect}$. Thus, the writer's identity belongs to the one who has the maximum output sum obtained as defined in (11):

$$Iw = \underset{k \in \{1,...,Nw\}}{\mathrm{ArgMax}} \left\{ \sum_{i=1}^{S} T_{affect}(X_i, k) \right\} \quad (11)$$

with:
$Iw$: index number of the identified writer.
$k$: index number of the $k^{th}$ suspect writer.
$Nw$: number of suspect writers.
$S$: number of segments of N Beta strokes recuperated from the tested handwriting data.
$T_{affect}$: rate of affectation of the $i^{th}$ tested segment to the $k^{th}$ writer, as defined in (12):

$$T_{affect}(X_i, k) = \left[ net_j(X_i) \right]_k \quad (12)$$

with:
- $Xi$: feature vector of the $i^{th}$ tested segment affected to the $j^{th}$ subgroup.
- $net_j$: output of the $j^{th}$ trained network to which the feature vector $X_i$ is assigned.

## 4 EXPERIMENTAL RESULTS AND DISCUSSION

We have conducted experiments on two publicly available datasets on Latin and Arabic scripts. These include IBM_UB_1 and ADAB datasets as discussed in the following.

### 4.1 Experiments on IBM_UB_1 dataset

IBM_UB_1 dataset [32], [35] is initially collected by IBM and later released by University at Buffalo. IBM_UB_1 dataset contains 6654 online cursive handwritten pages in English produced by 43 writers and divided into 4138 summary pages and 2516 query pages. In our experiments, we use some samples from the summary pages.

The different 43 writers do not contribute equal numbers of pages to the IBM_UB_1 dataset. In order to give the same chance to all writers and to investigate the effect of a growing amount of training data, we extracted 2 sub-datasets S1 and S2 from IBM_UB_1 dataset which contain respectively 6 and 12 pages for training per each writer (except 6 pages for writer 22 who contributed a small number of pages to the dataset). In the test step, we use 3 pages for each one of the 43 writers that conduct 3 sub-tests. Moreover, we rank the writers according to their obtained sum of output affectation rate as detailed in 3.5.

Accordingly, for each experiment, the total number of tests is: 3 x 43 = 129 tests. For example, in the experiment of the Extended Beta-Elliptic model from S2, we get 118 correct identifications and 11 false identifications, which gives an identification rate of 91.47%. The different experimental results on IBM_UB_1 dataset are detailed in Table 10.

TABLE 10
EXPERIMENTAL RESULTS ON IBM_UB_1 DATASET

| N | Features extraction technique | Identification rate for S1 (%) | | | Identification rate for S2 (%) | | |
|---|---|---|---|---|---|---|---|
| | | Top1 | Top5 | Top10 | Top1 | Top5 | Top10 |
| 1 | Beta-Elliptic model | 84.49 | 93.80 | 96.12 | 85.27 | 93.02 | 96.90 |
| 2 | Extended Beta-Elliptic model | 90.69 | 94.57 | 96.12 | 91.47 | 94.57 | 98.45 |
| 3 | Beta-Elliptic model + FEPC | 94.57 | 96.12 | 99.22 | 96.12 | 96.90 | 98.45 |
| **4** | **Extended Beta-Elliptic model + FEPC** | **96.12** | **98.45** | **99.22** | **96.90** | **100.00** | **100.00** |

From experimental results, we can see that: First, as Table 10 shows, the features extracted by the Extended Beta-Elliptic model describing simultaneously the static and the dynamic features are effective to discriminate the styles of handwriting and to identify the writer.

Second, it is important to observe that from the results, the combination of Fuzzy Elementary Perceptual Codes with Beta-Elliptic model and with Extended Beta-Elliptic model improves the identification rate. This indicate that the Fuzzy Elementary Perceptual Codes are robust description of handwriting to be used in writer identification.

Third, these experiments have shown that the results increase proportionally with the number of training pages. For the example of the combinaison between the Extended Beta-Elliptic model and the Fuzzy Elementary Perceptual Codes, the result was improved from 96.12% for S1 to 96.90% for S2.

Fourth, as shown in Figure 12, we plot the CMC Curves of performed writer identification systems on S2 from IBM_UB_1 dataset. The best identification rate is 96.90% in Top1 for the proposed system using the combinaison between the Extended Beta-Elliptic model and the Fuzzy Elementary Perceptual Codes. Only two writers were not well identified (writer 13 and writer 22). This can be explained that for the writer 22, we used a few pages in training as we said in the last sub-section.

At last, as Table 11 shows, our proposed system using the combinaison between the Extended Beta-Elliptic model and the Fuzzy Elementary Perceptual Codes achieves the best identification rate with 96.90% compared to the previous systems using IBM_UB_1 dataset [32], [34], [38] and [39].

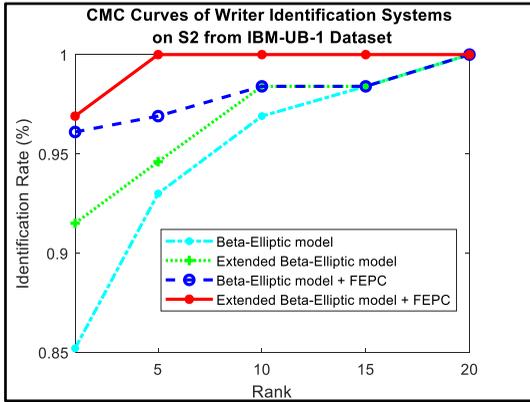

Fig. 12. CMC Curves of Writer Identification Systems on S2 from IBM_UB_1 Dataset

TABLE 11
COMPARISON OF IDENTIFICATION RATE WITH OTHER SYSTEMS USING IBM_UB_1 DATASET

| N | System | Identification rate (%) |
|---|---|---|
| 1 | System of Shivram *et al.* [32] | 89.47 |
| 2 | System of Shivram *et al.* [34] | 90.02 |
| 3 | System of Venugopal *et al.* [38] | 94.37 |
| 4 | System of Venugopal *et al.* [39] | 96.10 |
| 5 | **Proposed system using Extended Beta-Elliptic model + FEPC** | **96.90** |

### 4.2 Experiments on ADAB dataset

ADAB dataset consists of more than 33000 Arabic words, includes Tunisian town and village names, collected from 166 writers. It was developed in cooperation between the Institut fuer Nachrichtentechnik and REsearch Groups in Intelligent Machines [50], [51], [52] in order to advance the research and development of Arabic handwritten text [53], [54], [55], [56], [57], [58], [59].

To cope with the imbalanced number of words between writers in ADAB dataset, we selected sub-dataset of 19 writers, which contain the same words number for each writer. We used 67 words per writer in the training and 30 words per writer in the test divided into 3 sub-test contains each one 10 words as in the experiment done in [25]. Accordingly, for each experiment, the total number of tests is: 3 x 19 = 57 tests. For example, in the experiment of system using the Extended Beta-Elliptic model, we get 54 correct identifications and 3 false identifications, which gives an identification rate of 94.74%. The different experimental results on ADAB dataset are detailed in Table 12.

TABLE 12
EXPERIMENTAL RESULTS ON ADAB DATASET

| N | Features extraction technique | Identification rate (%) | | |
|---|---|---|---|---|
| | | Top1 | Top2 | Top3 |
| 1 | Beta-Elliptic model | 94.74 | 96.49 | 100.00 |
| 2 | Extended Beta-Elliptic model | 94.74 | 98.25 | 100.00 |
| 3 | Beta-Elliptic model + FEPC | 96.49 | 98.25 | 100.00 |
| 4 | **Extended Beta-Elliptic model + FEPC** | **98.25** | **100.00** | **100.00** |

From experimental results, we can see that firstly, the features extracted by the Extended Beta-Elliptic model are effective to discriminate the styles of handwriting and to identify the writer.

Secondly, these experiments show that the combination of Fuzzy Elementary Perceptual Codes with Beta-Elliptic model and with Extended Beta-Elliptic model improves the identification rate.

Thirdly, as shown in Figure 13, we plot the CMC Curves of performed writer identification systems on ADAB dataset. The combination between Extended Beta-Elliptic model and Fuzzy Elementary Perceptual Codes has yielded the best result with identification rate of 98.25%. In fact, we get 56 correct identifications and 1 false identifications.

TABLE 13
COMPARISON OF IDENTIFICATION RATE WITH OTHER SYSTEM USING ADAB DATASET

| N | System | Identification rate (%) |
|---|---|---|
| 1 | System of Dhieb *et al.* [25] | 91.22 |
| 2 | **Proposed system using Extended Beta-Elliptic model + FEPC** | **98.25** |

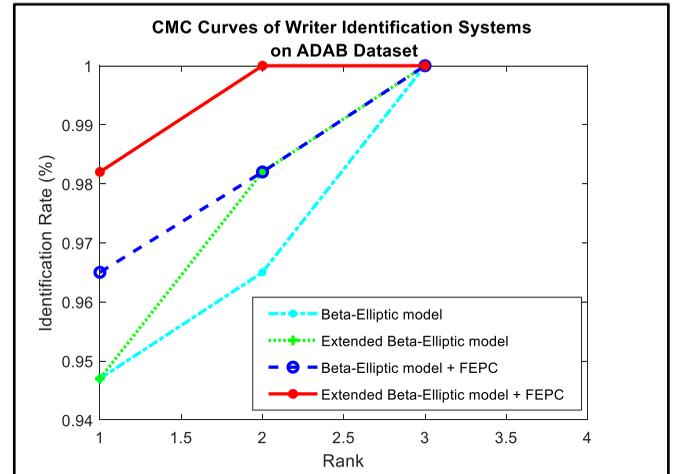

Fig. 13. CMC Curves of Writer Identification Systems on ADAB Dataset

As Table 13 shows, our proposed system using the combinaison between the Extended Beta-Elliptic model and the Fuzzy Elementary Perceptual Codes achieves the best identification rate compared to the system presented in [25] using ADAB dataset. Knowing that in [25], Dhieb *et al.* use the Beta-Elliptic model in feature extraction and Feed Forward Neural Networks in classification.

## 5 CONCLUSION

The originality of the work presented in this paper resides in the development of a new online text independent writer identification system based on the preprocessing and the segmentation of online handwriting into Beta strokes. We decompose the online handwriting into segments, which are classified into groups and subgroups. we define an extended version of Beta-Elliptic model and Fuzzy Elementary Perceptual Codes to characterize the handwriting of writers. Moreover, we use Deep Neural Network as a classifier. The latter includes the stacked sparse autoencoder in an unsupervised manner, followed by a softmax classifier that employs supervised learning to fine-tune the entire architecture.

Experimental results reveal that the proposed combination between the Extended Beta-Elliptic model and the Fuzzy Elementary Perceptual Codes achieves interesting results as compared to those of the existing writer identification systems using IBM_UB_1 and ADAB datasets. It proves that the features extraction technique from the proposed model and the combination with the Fuzzy Elementary Perceptual Codes is a powerful tool for the online writer identification on Latin and Arabic scripts.

Prospectively, we plan to test our proposed system with the Recurrent Neural Networks, and more precisely with Long Short Term Memory networks. In fact, the latter is attracting much attention for computer vision classification problems such as human actions recognition and multi-label image classification. In addition, we aim to use the combination between the Extended Beta-Elliptic model and the Fuzzy Elementary Perceptual Codes on other tasks like style classification, signature verification and handwriting recognition.

## ACKNOWLEDGMENT

The research leading to these results has received funding from the Ministry of Higher Education and Scientific Research of Tunisia under the grant agreement number LR11ES48.